\documentclass[letterpaper, 10 pt, conference]{ieeeconf}
\IEEEoverridecommandlockouts
\usepackage{float}
\usepackage{afterpage}
\usepackage{hyperref}
\usepackage{xspace}
\usepackage{graphicx} \graphicspath{{figures/}} 
\usepackage{amsmath,amssymb,nicefrac,bbm,pifont}
\usepackage{cite}
\usepackage{tcolorbox}
\usepackage[linesnumbered,ruled,vlined]{algorithm2e}
\usepackage{acronym}
\usepackage[skip=3pt,font=small]{subcaption}
\usepackage[skip=3pt,font=small]{caption}
\usepackage[dvipsnames,svgnames,x11names,table]{xcolor}
\usepackage[capitalise,noabbrev,nameinlink]{cleveref}
\usepackage{booktabs,tabularx,colortbl,multirow,multicol,array,makecell,tabularray}
\usepackage{overpic,wrapfig}
\usepackage[misc]{ifsym}
\usepackage{orcidlink}
\usepackage{dblfloatfix}
\usepackage{siunitx}
\usepackage{xr}
\usepackage{array}
\usepackage{graphicx}
\usepackage{dsfont}
\usepackage{diagbox}
\usepackage{multirow}

\makeatletter
\let\@origthebibliography\thebibliography
\def\thebibliography#1{\@origthebibliography{#1}%
  \linespread{0.88}\selectfont
  \setlength{\itemsep}{0pt}\setlength{\parsep}{0pt}\setlength{\parskip}{0pt}\interlinepenalty100}
\makeatother

\definecolor{sweepColor}{RGB}{154,170,167}
\definecolor{hammerColor}{RGB}{168,162,200}
\definecolor{torqueColor}{RGB}{205,179,165}
\definecolor{scoopColor}{RGB}{167,193,205}
\definecolor{gblue}{HTML}{4285F4}
\definecolor{gred}{HTML}{DB4437}
\definecolor{ggreen}{HTML}{0F9D58}
\definecolor{ggrey}{HTML}{E9E9E9}
\definecolor{lightpurple}{HTML}{d8c7e5}
\definecolor{darkpurple}{HTML}{b18dcc}
\definecolor{gbest}{HTML}{FFFFFF}
\definecolor{gsecond}{HTML}{FFFFFF}
\definecolor{pinkred}{HTML}{ea99a5}
\definecolor{pinkblue}{HTML}{96b4e2}
\definecolor{SkyBlue}{RGB}{135, 206, 235}

\crefname{algorithm}{Alg.}{Algs.}
\Crefname{algocf}{Algorithm}{Algorithms}
\crefname{section}{Sec.}{Secs.}
\Crefname{section}{Section}{Sections}
\crefname{table}{Tab.}{Tabs.}
\Crefname{table}{Table}{Tables}
\crefname{figure}{Fig.}{Figs.}
\Crefname{figure}{Figure}{Figures}
\crefname{equation}{Eq.}{Eqs.}
\Crefname{equation}{Equation}{Equations}
\crefname{appendix}{Appx.}{Appxs.}
\Crefname{appendix}{Appendix}{Appendices}

\makeatletter
\DeclareRobustCommand\onedot{\futurelet\@let@token\@onedot}
\def\@onedot{\ifx\@let@token.\else.\null\fi\xspace}
\def\eg{\textit{e.g}\onedot}

\def\vs{\textit{vs}\onedot}

\def\wrt{\textit{w.r.t}\onedot}

\makeatother

\DeclareMathOperator*{\argmax}{arg\,max}
\DeclareMathOperator*{\argmin}{arg\,min}

\newcommand{\frameworkName}{\textbf{\texttt{HOT}}}
\newcommand{\frameworkFullName}{\textbf{H}ierarchical \textbf{O}ptimization for \textbf{T}ool design}
\newcommand{\simName}{\texttt{DiffRedMax}\xspace}

\newcommand{\sweepTask}{\texttt{\textbf{SweepBalls}}\xspace}
\newcommand{\torqueTask}{\texttt{\textbf{TorqueBolt}}\xspace}
\newcommand{\scoopTask}{\texttt{\textbf{ScoopBalls}}\xspace}
\newcommand{\hammerTask}{\texttt{\textbf{Hammer\&ExtractNail}}\xspace}
\newcommand{\exploredMetricName}{FSI\xspace}
\newcommand{\exploredMetricFullName}{First-Success Index\xspace}

\newcommand{\bassName}{\textbf{\texttt{\acs{bass}}}}

\newcommand{\bassFullAnnotAbbr}{\textbf{B}ehavior-\textbf{A}ware \textbf{S}tructural \textbf{S}earch (\textbf{\texttt{\acs{bass}}})}

\acrodef{ai}[AI]{Artificial Intelligence}
\acrodef{vlm}[VLM]{Vision Language Model}
\acrodef{llm}[LLM]{Large Language Model}
\acrodef{bass}[BASS]{Behavior-Aware Structural Search}
\acrodef{dag}[DAG]{directed acyclic graph}

\title{\LARGE \bf Robot Tool Design from Scratch\\via Behavior-Aware Hierarchical Optimization}

\author{Yinghan Chen$^{1,2,3,4,5*}$, Xiyao Tian$^{1,2,3,4,5*}$, Yizan Dai$^{1,2,3,4,5}$, Yuyang Li$^{1,2,4,5,\dagger}$, and Yixin Zhu$^{2,1,4,5,6,7,\dagger}$ \\
\textcolor{magenta}{\textbf{\texttt{https://hot.yinghanchen.com}}} \\
\thanks{$^*$ Equal contributors. $^\dagger$ Corresponding author.}
\thanks{$^1$ Institute for AI, Peking University.}
\thanks{$^2$ School of Psychological and Cognitive Sciences, Peking University.}
\thanks{$^3$ Yuanpei College, Peking University.}
\thanks{$^4$ Beijing Key Laboratory of Brain-Computer Interface and Mental Health Modulation, Peking University.}
\thanks{$^5$ State Key Laboratory for General AI, Peking University.}
\thanks{$^6$ Embodied Intelligence Lab, PKU-Wuhan Institute for AI.}
\thanks{$^7$ XS Vision.}
}

\begin{document}

\let\oldtwocolumn\twocolumn
\renewcommand\twocolumn[1][]{
    \oldtwocolumn[{#1}{
        \vspace{-30pt}
        \centering
        \includegraphics[width=\linewidth]{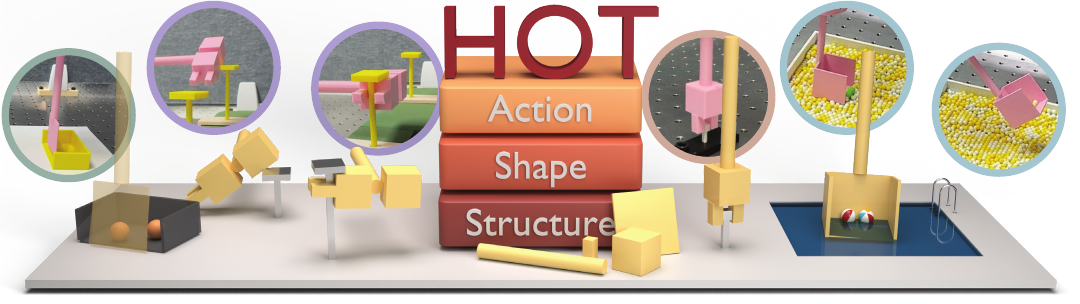}
        \captionof{figure}{
        \textbf{\frameworkName{} designs robot tools from scratch by jointly optimizing structure, shape, and action.}
        Tools are composed from primitive assets without a prescribed structure template, with structural search and physical optimization in a closed loop.
        \textcolor{sweepColor}{\textbf{(a)~\sweepTask:}} sweep two balls into a dustpan.
        \textcolor{hammerColor}{\textbf{(b)~\hammerTask:}} drive one nail into a board and then pry out a second.
        \textcolor{torqueColor}{\textbf{(c)~\torqueTask:}} engage a bolt and turn it to a target angle.
        \textcolor{scoopColor}{\textbf{(d)~\scoopTask:}} scoop two balls floating on water and lift them clear.
        }
        \label{fig:teaser}
    }]                  
}

\maketitle

\thispagestyle{empty}
\pagestyle{empty}

\begin{abstract}

The ability to design a tool for a task marks a level of intelligence beyond merely understanding, selecting, or using one~\cite{vaesen2012cognitive,qin2023robot}.
Existing methods for robotic tool design typically optimize a tool's continuous shape and action~\cite{xu2021end,li2023learning} within a structure that is prescribed or generated beforehand~\cite{nair2019autonomous,lin2025robotsmith}, so the structure itself stays outside the physical optimization loop.
We study task-driven tool design from scratch, where tool structure, shape, and action are all derived from the desired physical outcome.
Here we show that the three elements can be designed jointly by \frameworkName{}, a hierarchical optimization whose upper level searches over discrete tool structures with \bassName{}, while lower-level physical optimization evaluates their task behavior and returns milestone progress as behavioral evidence for the search, ultimately providing jointly optimized shape and action.
On four tool-use tasks with distinct physical functions, \frameworkName{} discovers functional structures after evaluating only a small fraction of search spaces containing up to 56 million structures, and the subsequent refinement of their geometry lowers the task loss on all tasks while preserving success, through deformations that are functionally interpretable.
Once 3D printed, the tools accomplish all tasks on a real robot with the actions found in simulation.
Designing tools from required physical effects, rather than a catalog of known tools, is a step toward the open-ended tool making seen in humans and animals~\cite{muller2022stone,bayern2018compound}.
\end{abstract}

\section{Introduction}\label{sec:intro}

Physical tools let humans and robots produce physical effects beyond the reach of their bodies~\cite{zhu2020dark,muller2022stone,billard2019trends,vaesen2012cognitive,johnson2004neural,allen2020rapid,qin2023robot}. Prior work studies how tools function through animal behavior and computational models~\cite{allen2020rapid,stoytchev2005behavior,mar2018can,turpin2021gift,vaesen2012cognitive}, how to select one from candidates~\cite{zhu2015understanding}, how to use one~\cite{zhu2015understanding,qin2020keto,zorina2022learning,qi2024learning,tang2025mimicfunc,chen2025tool,kedia2026simtoolreal}, and typically assumes that a suitable tool with an appropriate structure and shape already exists. Tool design asks a harder question: given a desired physical outcome, can a robot design and use a suitable tool from scratch~\cite{zhu2015understanding,zhu2020dark,nair2019autonomous,liu2023learning,lin2025robotsmith}?

Human and animal studies suggest what such design involves. New Caledonian crows assemble short components into compound tools~\cite{bayern2018compound} and craft hooks by selecting, trimming, and bending raw material~\cite{weir2002shaping,hunt2004crafting}; human stone-tool production alters morphology through successive shaping~\cite{muller2022stone}. These examples expose three coupled dimensions of a tool: its \textit{structure}, how components compose into a functional assembly; its \textit{shape}, the continuous geometry of those components; and its \textit{action}, how the tool establishes contact and produces the desired effect. The three cannot be designed independently: a concave structure that scoops well may be useless for exerting impact, and a favorable shape helps only if a feasible action can establish the required interaction. Tool functionality is therefore task-dependent and emerges jointly~\cite{zhu2015understanding,billard2019trends,zhu2020dark} from structure, shape, action, and their physical interaction with the environment.

Existing work covers this design space only in part. Tool selection operates over predefined candidates~\cite{zhu2015understanding,zhu2026generalizable}, and tool optimization refines continuous geometry under a fixed structure~\cite{xu2021end,li2023learning}. Recent methods generate a structure before optimizing its shape or use~\cite{nair2019autonomous,lin2025robotsmith}, but the structure, once generated, stays outside the physical optimization loop: the outcome of the optimized shape and action does not inform how the structure itself should evolve. What is missing is a unified framework that determines structure, shape, and action jointly from the physical effects a task requires, which is what defining a tool completely from scratch entails.

We propose \frameworkName{} (\frameworkFullName{}), which designs tool structure, shape, and action from a task's desired physical outcome (\cref{fig:teaser}). We formulate tool design as a bilevel optimization: the upper level searches over discrete structures, while the lower level optimizes the shape and action of each candidate to accomplish the task. The task performance is fed back to the upper level, closing the loop between structural search and physical optimization.

The upper level is challenging as successful tools are sparse in a vast structural space, and each evaluation requires expensive lower-level optimization in simulation. We therefore introduce \bassFullAnnotAbbr{}, which constructs structures with a grammar and uses the milestone-based progress of evaluated tools to update a posterior over each partial construction state's likelihood of yielding a successful tool. Bayesian lookahead then concentrates evaluations on promising structural regions. Unlike candidates in Bayesian optimization or states in game-tree search, partial structures have no task value themselves; \bassName{} is designed for this setting where evidence arises only from optimized complete descendants. This behavior-aware feedback enables \frameworkName{} to jointly determine which \textit{structure} to build, how to refine its \textit{shape}, and which \textit{actions} to use.

We evaluate \frameworkName{} on four manipulation tasks with distinct physical effects and structural requirements. Each task is specified by milestones and losses on the desired interaction outcomes, without a tool identity or a structural template.  \frameworkName{} designs tools and actions that succeed in simulation and on a real robot, and refining the discovered structures lowers the task loss on all tasks while preserving success. The search can yield structurally different solutions for the same task, as well as multifunctional structures whose functional regions serve distinct physical functions within a single tool.

Our contributions are three-fold: (i)~\frameworkName{}, a task-driven bilevel paradigm that designs tools from scratch through discrete structural search and continuous shape-action optimization; (ii)~\bassName{}, a behavior-aware search strategy for combinatorial structural design problems like tool design, whose partial states must be evaluated with completed descendants; and (iii)~a task suite that evaluates task-driven tool design, where \frameworkName{} is validated in simulation and real world.

\section{Related Work}

\paragraph*{Tool Use and Design for Robotics}
Most robotic tool research works with existing tools: it studies tool functionality~\cite{stoytchev2005behavior,mar2017self,do2018affordancenet,myers2015affordance,turpin2021gift,qin2020keto}, selects task-appropriate tools from predefined candidates~\cite{zhu2015understanding,zhu2026generalizable,abelha2016model,xie2019improvisation}, and performs tool use with data-driven~\cite{qi2024learning,chen2025tool,tang2025mimicfunc,kedia2026simtoolreal,xiong2025ag2x2} or model-based methods~\cite{fitzgerald2021modeling,holladay2019force,toussaint2018differentiable}, generally assuming a predefined tool structure and shape. Recent work extends the problem to tool design: some methods optimize continuous geometry under a fixed structure~\cite{xu2021end,li2023learning,liu2023learning}, others construct tools by assembling parts~\cite{nair2019tool,nair2019autonomous,zlokapa2022integrated} or generate structures with \acp{vlm}~\cite{lin2025robotsmith}. Each covers the structure-shape-action space of \cref{sec:intro} only in part: the structure is either fixed in advance or generated before the physical optimization, so the outcome of that optimization does not feed back into the structural choice. \frameworkName{} connects the two stages through \bassName{}, guiding the structural search with optimized task performance of evaluated tools.

\paragraph*{Learning from Experience}
Many learning and search paradigms improve future decisions with past outcomes. Q-learning propagates downstream returns to action values~\cite{watkins1992q}; Monte Carlo and Bayesian tree search guide exploration with descendant outcomes~\cite{kocsis2006bandit,tesauro2010bayesian}; and Bayesian active search and optimization allocate expensive evaluations through posterior beliefs, including nonmyopic, tree-structured, and combinatorial settings~\cite{garnett2012bayesian,jiang2017efficient,jenatton2017bayesian,baptista2018bayesian,wu2019practical}. In these paradigms, outcomes attach to decision states, search states, or candidates that are themselves evaluable. Our setting differs: a partial tool structure has no physical task value, and evidence arrives only by completing a descendant and optimizing its use. \bassName{} maintains a posterior over each partial structural region from such outcomes and applies a Bayesian lookahead to decide where to evaluate next.

\paragraph*{Differentiable Simulation for Design and Control}
Differentiable simulation provides gradients of simulated dynamics \wrt control inputs or physical parameters, enabling gradient-based optimization through physical interaction. Existing approaches address non-smooth contact by differentiating complementarity-based contact models or their smooth relaxations~\cite{howell2022dojo,werling2021fast,le2023single}, and some support implicit shape representations~\cite{strecke2021diffsdfsim,le2023differentiable}; DiffTaichi and Warp provide automatic differentiation infrastructure~\cite{hu2020difftaichi,miles2022warp}, while Brax and Newton emphasize parallel simulation on accelerators~\cite{freeman2021brax,nvidia2025newton}. Such gradients enable policy learning through simulated trajectories~\cite{xu2022accelerated}. We adopt \simName~\cite{xu2021end}, a differentiable extension of RedMax~\cite{wang2019redmax}, because it provides gradients \wrt both the actions and the continuous shape parameters of a rigid assembly, which is exactly what the lower-level optimization requires.

\begin{figure*}[t!]
    \centering
    \includegraphics[width=\linewidth]{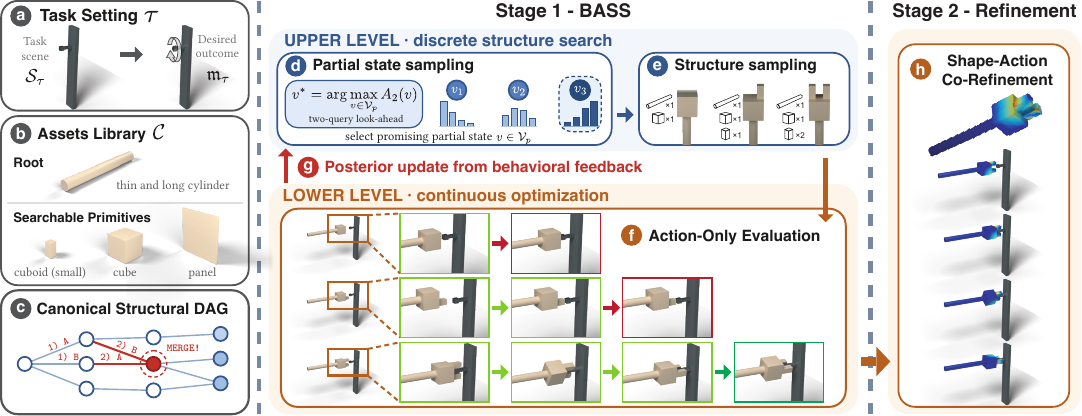}
    \caption{\textbf{The \frameworkName{} framework.} (a)~A task specifies a scene and a desired physical outcome. (b)~Tools are composed by a constructive grammar from a fixed handle and a library of searchable primitives. (c)~Equivalent construction states are canonicalized into a \ac{dag} that aggregates their behavioral evidence. In Stage~1, \bassName{} selects a promising partial state by Bayesian lookahead (d), samples a complete structure (e), evaluates it under action-only optimization (f), and propagates the achieved milestones back as behavioral feedback (g). Stage~2 co-refines the shape and action of each retained structure (h).}
    \label{fig:framework}
\end{figure*}

\section{Tool Design with \frameworkName{}}\label{sec:method}

We consider a tool-use task $\tau$ defined by a physical scene $\mathcal{S}_{\tau}$ and a desired final outcome $\mathfrak{m}_{\tau}$: $\mathcal{S}_{\tau}$ specifies the task objects, their initial states, and physical settings; $\mathfrak{m}_{\tau}$ specifies the completion condition (\eg, a bolt reaching a target rotation in \cref{fig:framework}(a)). Computationally, we characterize the task as $\tau=\left(\mathcal{S}_{\tau},\mathcal{M}_\tau,\mathcal{L}_\tau\right)$. $\mathcal{M}_\tau=(\mathfrak{m}_1,\dots,\mathfrak{m}_m)$ is a sequence of $m\geq 1$ ordered milestones ending at $\mathfrak{m}_m=\mathfrak{m}_\tau$: $m=1$ specifies only the final outcome, while $m>1$ adds intermediate physical conditions without prescribing a particular tool or action sequence. For the bolt-torquing example, they may include approaching, contacting, and rotating the bolt to a target angle. A differentiable loss $\mathcal{L}_{\tau}$ over these desired states provides a continuous objective, serving as a surrogate for maximizing achieved milestones for a given structure $T$. Given $\tau$, our goal is to find a suitable tool structure and shape, with actions that achieve $\mathfrak{m}_{\tau}$.

We formulate tool design as a bilevel optimization (\cref{sec:formulation}) and introduce a practical two-stage approximation (\cref{sec:two_stage}). We then describe the structural representation (\cref{sec:grammar}), the lower-level optimization (\cref{sec:lower}), and the shape-action co-refinement (\cref{sec:refine}); \cref{sec:bass} presents \bassName{}, the structural search that drives Stage~1.

\subsection{Tool Design as Bilevel Optimization}\label{sec:formulation}

We formulate the problem over three coupled spaces: tool structure, shape, and action. Let $\mathcal{T}$ denote the structural space admissible under the construction rules of \cref{sec:grammar}, $\Psi(T)$ the shape parameter space of a structure $T\in\mathcal{T}$, and $\mathcal{U}$ the space of valid controls. A candidate tool with its use is $\mathfrak{t}=\langle T,\psi,u\rangle$: $T$ is a tree-structured rigid assembly whose nodes are primitive components and whose edges encode their attachment choices; $\psi\in\Psi(T)$ describes the continuous geometry of the primitives and their attachments; and $u\in\mathcal{U}$ is the control sequence that manipulates the assembled tool. Given a tool design $\mathfrak{t}$, we simulate its tool use to obtain the task loss $\mathcal{L}_\tau(T,\psi,u;\mathcal{S}_\tau)$ and the ordered set of achieved milestones, $\mathcal{M}_\mathfrak{t}:=M_\tau(T,\psi,u;\mathcal{S}_\tau)\subseteq\mathcal{M}_\tau$. Tool design is then a discrete-continuous bilevel problem: the upper level seeks the structure that achieves the most milestones with the shape and action from a lower-level continuous optimization,
\begin{equation}
    \begin{aligned}
        &T^{*}=\argmax_{T\in\mathcal{T}}\;\vert M_\tau(T,\psi_T^{*},u_T^{*};\mathcal{S}_\tau)\vert\\
        &\text{subject to }\left(\psi_T^{*},u_T^{*}\right)=\argmin_{\psi\in\Psi(T),\,u\in\mathcal{U}}\;\mathcal{L}_{\tau}(T,\psi,u;\mathcal{S}_\tau).
    \end{aligned}
    \label{eq:bilevel}
\end{equation}

\subsection{A Two-Stage Approximation}\label{sec:two_stage}

Solving \cref{eq:bilevel} directly is expensive: it requires a joint shape-action optimization for every structure candidate, most of which cannot accomplish the task. We therefore approximate it in two stages. Stage~1, the loop in \cref{fig:framework}\mbox{(d--g)}, searches for promising structures under action-only optimization: each candidate $T$ is instantiated in simulation with its primitives at their original library shapes, $\psi_0(T)$, and only its action is optimized with the shape frozen,
\begin{equation}
    u_T^\dagger=\argmin_{u\in\mathcal{U}}\;\mathcal{L}_{\tau}\left(T,\psi_0(T),u;\mathcal{S}_\tau\right),
    \label{eq:stage1}
\end{equation}
which keeps each structure evaluation cheap relative to the joint problem. The milestones $\mathcal{M}_{\mathfrak{t}}$ achieved by the optimized rollout provide task-progress evidence for \bassName{} (\cref{sec:bass}), which returns a set $\mathcal{T}^\dagger$ of retained structures that complete the task, each with its optimized action $u_{T^\dagger}^\dagger$. Stage~2 (\cref{fig:framework}h) then refines shape and action jointly for each $T^\dagger\in\mathcal{T}^\dagger$:
\begin{equation}
    \left(\psi_{T^\dagger}^*,u_{T^\dagger}^*\right)=\argmin_{\psi\in\Psi(T^\dagger),\,u\in\mathcal{U}}\;\mathcal{L}_{\tau}(T^\dagger,\psi,u;\mathcal{S}_\tau),
    \label{eq:stage2}
\end{equation}
with initialization $(\psi_0(T^\dagger),u_{T^\dagger}^\dagger)$. This finally yields the tool design with optimal performance: $\mathfrak{t}^*=\langle T^*,\psi_{T^*}^*,u_{T^*}^*\rangle.$

\subsection{Structural Representation with Constructive Grammar}\label{sec:grammar}

A structure is built by selecting and connecting primitives from a compact asset library $\mathcal{C}$ (\cref{fig:framework}b) into a tree $T$, which yields diverse structures with parameterized shapes from a few assets. Each node is a primitive, and each edge connects a parent and a child through a pair of docks. A dock is a local frame anchored on a face of a primitive with its $z$ axis along the face normal; two primitives are connected by aligning their dock anchors with opposing outward normals at one of four in-plane orientations, $\phi=k\pi/2$ with $k\in\{0,1,2,3\}$.

As different regions of a tool may serve different physical roles, treating the whole structure as one task-facing unit is too coarse to distinguish localized functions or to support multifunctional designs. We therefore introduce functional groups: each group is a subtree that represents one functional region; groups cannot be nested, and each primitive belongs to at most one group. Each node carries a binary label $z\in\{0,1\}$: $z=1$ marks the node as the root of a group consisting of its subtree. Each group is matched to a task-specific loss term when introduced by the grammar, and this assignment is retained as part of the construction state.

Each structure is generated by a depth-first construction sequence. Let $s$ denote the current construction state, namely the partial assembly together with its active primitive. Starting from the root primitive marked as active, the grammar repeatedly applies one of the admissible operations
\begin{equation}
    \mathcal{O}(s)=\left\{\texttt{AddLink}(c,d_p,d_c,\phi,z),\ \texttt{End}\right\}.
\end{equation}
\texttt{AddLink} attaches primitive $c\in\mathcal{C}$ to the active primitive through parent dock $d_p$ and child dock $d_c$ with rotation $\phi$; $z$ marks whether $c$ starts a functional group consisting of the subtree rooted at it; $c$ then becomes the active primitive. \texttt{End} terminates the current branch and backtracks to the parent, marking it as active; \texttt{End} at the root terminates the construction to yield a complete structure $T$. Invalid operations (against dock compatibility, component-count bounds, functional-group requirements, or self-collision) are excluded from $\mathcal{O}(s)$. In practice, the root is a cylindrical handle for grasping or mounting, and $\mathcal{C}$ contains three cuboids of different dimensions to form functional geometries (\cref{fig:framework}b).

\subsection{Lower-Level Optimization in Differentiable Simulation}\label{sec:lower}

Lower-level optimization runs in \simName~\cite{wang2019redmax,xu2021end}. For a structure $T$, we instantiate the scene $\mathcal{S}_{\tau}$ and assemble the tool by connecting its primitives with fixed joints. In Stage~1, the simulator provides gradients of $\mathcal{L}_{\tau}$ \wrt action for optimizing $u$ using a task-stage-wise trust-region method, with initial shapes $\psi=\psi_0(T)$ fixed. In Stage~2, we alternate updates on shapes and actions (\cref{sec:refine}). $\mathcal{L}_{\tau}$ combines four kinds of terms,
\begin{equation}
    \mathcal{L}_{\tau}
    = \lambda_{\tau}^{\mathrm{out}} \mathcal{L}_{\tau}^{\mathrm{out}}
    + \lambda_{\tau}^{\mathrm{geo}} \mathcal{L}_{\tau}^{\mathrm{geo}}
    + \lambda_{\tau}^{\mathrm{int}} \mathcal{L}_{\tau}^{\mathrm{int}}
    + \lambda_{\tau}^{\mathrm{reg}} \mathcal{L}_{\tau}^{\mathrm{reg}}.
\end{equation}
The outcome term $\mathcal{L}_{\tau}^{\mathrm{out}}$ measures the discrepancy between the current and target states (\eg, position, height, rotation) of the objects, independently of the tool; the geometry term $\mathcal{L}_{\tau}^{\mathrm{geo}}$ measures spatial relations (\eg, relative position, distance, clearance) among the tool's functional groups, the objects, and the scene; the interaction term $\mathcal{L}_{\tau}^{\mathrm{int}}$ encourages the desired physical interactions (\eg, contact forces or impulses); and the regularization term $\mathcal{L}_{\tau}^{\mathrm{reg}}$ penalizes action magnitude and temporal variation for efficient, smooth trajectories. The weights $\lambda_{\tau}^{(\cdot)}$ balance the terms.

\subsection{Shape-Action Co-Refinement}\label{sec:refine}

Stage~2 realizes \cref{eq:stage2} for a retained structure $T$ by keeping $T$ fixed and alternating shape perturbation with action re-optimization, starting from $(\psi_0(T),u_T^\dagger)$. Each iteration samples perturbation directions and progressively increases their magnitude to deform all eligible primitives. For every geometrically plausible proposal $\psi^{(j)}$ (filtered by penetration and connectivity), the action is re-optimized from the last accepted one, so that the trajectory adapts to the new shape. The pair $(\psi^{(j)},u^{(j)})$ is accepted if its rollout remains task-successful and its loss exceeds the Stage~1 loss by at most $\epsilon_s=2\%$, a tolerance that permits exploration; the final output must attain a strictly lower loss than the Stage~1 solution. A rejected proposal reduces the perturbation magnitude, and refinement continues from the last accepted pair, which also initializes the next perturbation step.

\section{Behavior-Aware Structural Search (\bassName{})}\label{sec:bass}

Unfolding the constructive grammar yields a discrete space far too large for exhaustive search, while each evaluation requires a lower-level optimization. \bassName{} thus guides Stage~1 with the physical outcomes of these optimizations. A partial construction state is not itself a structure, but defines the complete structures reachable through valid continuations. We first canonicalize equivalent construction states into a \ac{dag} (\cref{sec:dag}), estimate each partial state's task potential from the milestone progress of its evaluated complete descendants (\cref{sec:feedback}), and allocate following evaluations with Bayesian lookahead (\cref{sec:lookahead}).

\begin{figure*}[t!]
    \centering
    \includegraphics[width=\linewidth]{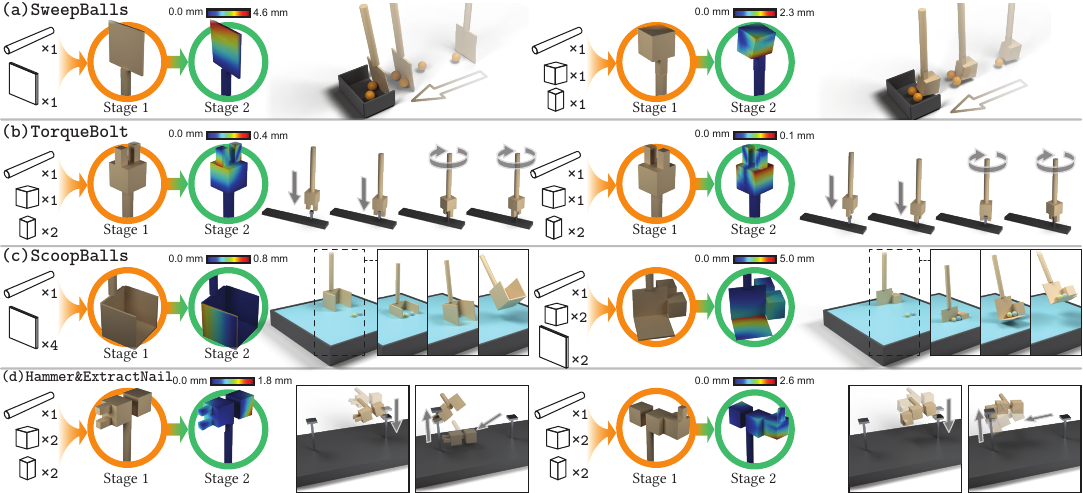}
    \caption{\textbf{Tools designed by \frameworkName{} and their use in simulation.} Each row shows two representative successful tool designs.
    For each design, we list the primitives selected by \bassName{} with their counts, the Stage~1 structure they compose, and its Stage~2 refinement, where the heat map shows the absolute vertex displacement from the Stage~1 shape in millimeters; the frames on the right show the refined tool rolled out on its task. Rows (a)--(d) correspond to \sweepTask, \torqueTask, \scoopTask, and \hammerTask.}
    \label{fig:sim}
\end{figure*}

\subsection{Search on a Canonical Structural DAG}\label{sec:dag}

In the search tree induced by the constructive grammar, different construction sequences can reach the same construction state (\cref{fig:framework}c). Searching such duplicates not only introduces redundant exploration but also separates observations that should provide evidence for the same states. We therefore canonicalize equivalent states into a \ac{dag}: each node $v$ is a canonical construction state of \cref{sec:grammar}, characterized by the current assembly, the active primitive, and the functional labels together with their assigned task functions; grammar operations induce the directed edges, and complete states form the terminal nodes.

\begin{algorithm}[t!]
    \caption{\bassName{} structural search}
    \label{alg:bass}
    \small
    \KwIn{Canonical \acs{dag}, task $\tau$, evaluation budget $B$}
    \KwOut{Retained structures $\mathcal{T}^\dagger$ with optimized actions}

    $\mathcal{V}_t\gets\varnothing$;\ 
    $\mathcal{V}_p\gets\{v_{\mathrm{root}}\}$;\ 
    $\mathcal{T}^\dagger\gets\varnothing$;\ 
    initialize $\mathcal{D}_{v_{\mathrm{root}}}$ and posterior\;

    \For{$b\gets1$ \KwTo $B$}{
        $A_2(w)\gets$\cref{eq:lookahead},$\quad \forall w\in\mathcal{V}_p$\;
        $v\gets\argmax_{w\in\mathcal{V}_p}A_2(w)$\;

        Sample $v=v^{(0)}\to\cdots\to v^{(n)}$ s.t.
        $v^{(n)}\notin\mathcal{V}_t$\;

        $T\gets T(v^{(n)})$\;
        $u_T^\dagger\gets\texttt{OptimizeAction}(T)$ by \cref{eq:stage1}\;
        $c\gets\left|M_\tau(T,\psi_0(T),u_T^\dagger;\mathcal{S}_\tau)\right|$\;

        $\mathcal{V}_t\gets\mathcal{V}_t\cup\{v^{(n)}\}$\;
        \lIf{$c=m$}{
            $\mathcal{T}^\dagger\gets\mathcal{T}^\dagger\cup\{T\}$
        }

        \ForEach{$w\in\{v^{(0)},v^{(1)},\ldots,v^{(n-1)}\}$}{
            \lIf{$w\notin\mathcal{V}_p$}{
                $\mathcal{V}_p\gets\mathcal{V}_p\cup\{w\}$; initialize posterior
            }
            $\mathcal{D}_w\gets\mathcal{D}_w\uplus\{c\}$;\ 
            update $\{\alpha_{w,\ell},\beta_{w,\ell}\}_{\ell=0}^{m-1}$ by \cref{eq:posterior}\;
        }

        $\mathcal{V}_p\gets
        \{w\in\mathcal{V}_p:\exists\text{ terminal }v\notin\mathcal{V}_t,\ w\leadsto v\}$\;
    }
\end{algorithm}

\cref{alg:bass} searches this \ac{dag} with an evaluated terminal set $\mathcal{V}_t$ and a frontier $\mathcal{V}_p$ of partial states that still have at least one unevaluated terminal descendant. Starting from the root state, each iteration selects a partial node $v\in\mathcal{V}_p$ (L3-4), samples a continuation $(v,v^{(1)},\ldots,v^{(n)})$ to an unevaluated terminal node $v^{(n)}\notin\mathcal{V}_\mathrm{t}$ (L5), evaluates the corresponding tool by the Stage~1 optimization (L6-8), records the terminal node and, if the tool completes the task, retains the structure (L9-10), and propagates the outcome to the traversed partial states (L11-13), adding the partial states along this sampled continuation to $\mathcal{V}_p$. Rather than choosing $v$ uniformly, we treat the behavioral outcome of each sampled completion as evidence for the task potential of the partial states traversed along its sampled continuation. \cref{sec:feedback} defines this estimate, and \cref{sec:lookahead} uses it to allocate the next evaluation.

\subsection{Behavioral Feedback on Partial States}\label{sec:feedback}

Most sampled structures fail, but differ substantially in how far their optimized behavior progresses through the milestones. We take the achieved milestone count $C=|\mathcal{M}_{\mathfrak{t}}|\in\{0,\ldots,m\}$ as the behavioral outcome of an evaluated tool $\mathfrak{t}$ ($C=m$ is task completion), and propagate it to partial states on the sampled continuation (\cref{alg:bass}, L11-13).

For each partial node $v$ we keep the multiset $\mathcal{D}_v=\{C_i\}_{i=1}^{n_v}$ of outcomes from sampled completions whose continuation traverses $v$; canonicalization ensures that observations accumulated when equivalent states are traversed contribute to the same $\mathcal{D}_v$. We use $\mathcal{D}_v$ to estimate how likely a new completion from $v$ is to progress through the remaining milestones. For milestone $\mathfrak{m}_\ell$, let
\begin{equation}
    h_{v,\ell}=\Pr\left(C=\ell\,\middle|\,C\geq\ell,\,v\right)
\end{equation}
be the probability that a new completion from $v$ stops at $\mathfrak{m}_\ell$; its complement is the probability of progressing beyond $\mathfrak{m}_\ell$. Before any observation, all partial nodes share the initial prior $h_{v,\ell}\sim\operatorname{Beta}(\alpha_\ell^0,\beta_\ell^0)$. Rather than specifying the Beta parameters directly, we parameterize the prior by the belief $\bar q_\ell$ that a completion progresses beyond $\mathfrak{m}_\ell$, estimated from random calibration rollouts, together with a prescribed prior strength $\kappa_\ell$, so that $\alpha_\ell^0=\kappa_\ell(1-\bar q_\ell)$ and $\beta_\ell^0=\kappa_\ell\bar q_\ell$. Given $\mathcal{D}_v$, the conjugate posterior parameters are
\begin{equation}
    \resizebox{0.91\linewidth}{!}{$\displaystyle%
        \alpha_{v,\ell}=\alpha_{\ell}^{0}+\sum_{i=1}^{n_v}\mathbf{1}[C_i=\ell],\quad
        \beta_{v,\ell}=\beta_{\ell}^{0}+\sum_{i=1}^{n_v}\mathbf{1}[C_i>\ell].
        \label{eq:posterior}
    $}
\end{equation}
An outcome $C_i=c<m$ thus contributes continuation evidence at every $\ell<c$ and stopping evidence at $\ell=c$, whereas a successful outcome $C_i=m$ contributes continuation evidence at all levels. Since the stopping probabilities $h_{v,\ell}$ are modeled independently across milestones, their posteriors remain independent. Therefore, the posterior mean continuation probability at $\mathfrak{m}_\ell$ and the resulting estimate of the success probability of the structural region below $v$ are
\begin{equation}
    \resizebox{0.91\linewidth}{!}{$\displaystyle%
        q_{v,\ell}=\frac{\beta_{v,\ell}}{\alpha_{v,\ell}+\beta_{v,\ell}},\quad
        \mu_v=\Pr\left(C=m\mid\mathcal{D}_v\right)=\prod_{\ell=0}^{m-1}q_{v,\ell}.
    $}
\end{equation}
$\mu_v$ is the task potential of $v$: the probability that a new complete descendant of $v$ succeeds after Stage~1 optimization.

\subsection{Bayesian Lookahead Allocation}\label{sec:lookahead}

Selecting $v=\argmax_{w\in\mathcal{V}_p}\mu_w$ would exploit the currently most promising state but ignore how the new observation changes later decisions. We instead use an $N$-query Bayesian lookahead over the current frontier $\mathcal{V}_p$. For $c<m$, the posterior predictive probability that a completion from $v$ achieves exactly $c$ milestones is $\pi_{v,c}=(1-q_{v,c})\prod_{\ell<c}q_{v,\ell}$, and $\pi_{v,m}=\mu_v$. Define the current search state,
\begin{equation}
    \Theta=\left(\alpha_{v,\ell},\beta_{v,\ell}\right)_{v\in\mathcal{V}_p,\ 0\leq\ell<m},
\end{equation}
which consists of the posterior parameters of all frontier states and milestone levels, and define $\Theta^{v,c}$ as the hypothetical state after observing outcome $c$ at $v$. The optimal probability of finding at least one successful structure within a remaining budget of $N$ queries is defined recursively by
\begin{equation}
    J_1(\Theta)=\max_{v\in\mathcal{V}_p}\mu_v(\Theta)
\end{equation}
and, for $N>1$,
\begin{equation}
    \begin{aligned}
        A_N(v;\Theta)&=\mu_v(\Theta)+\sum_{c=0}^{m-1}\pi_{v,c}(\Theta)\,J_{N-1}\left(\Theta^{v,c}\right),\\
        J_N(\Theta)&=\max_{v\in\mathcal{V}_p}A_N(v;\Theta).
    \end{aligned}
\end{equation}
The first term of $A_N$ is the success probability of the current query; the second is the optimal remaining search after each unsuccessful outcome and its posterior update. A larger $N$ anticipates more future evaluations, but the number of hypothetical outcomes grows rapidly with the horizon. We use $N=2$, accounting for one subsequent search decision while keeping the acquisition cheap to evaluate. During the lookahead, the frontier is held fixed and its posteriors are treated as independent, so a hypothetical observation at $v$ updates only the posterior of $v$. The acquisition is then
\begin{equation}
    A_2(v)=\mu_v+\sum_{c=0}^{m-1}\pi_{v,c}\max\left(\mu_v^{(c)},\max_{w\in\mathcal{V}_p\setminus\{v\}}\mu_w\right),
    \label{eq:lookahead}
\end{equation}
where $\mu_v^{(c)}$ is the success probability of $v$ after hypothetically observing $c$, and the next structural region is $v^*=\argmax_{v\in\mathcal{V}_p}A_2(v)$ (\cref{alg:bass}, L3-4).

\section{Experiments and Results}

\begin{table}[!t]
    \centering
    \footnotesize
    \setlength{\tabcolsep}{4pt}
    \caption{\textbf{Stage~1 structural search cost:} Design space size (terminal states in DAG), \exploredMetricName of \bassName{}, and time consumed (Core-h) for each task.}
    \label{tab:stage1}
    \resizebox{\linewidth}{!} {
        \begin{tabular}{lrrr}
            \toprule
            \textbf{Task} & \textbf{Design Space Size} & \textbf{\exploredMetricName} & \textbf{Core-h} \\
            \midrule
            \sweepTask  & 1,296,198  & 10,142 (7.82\textperthousand) & 916.63 \\
            \torqueTask & 1,296,198  & 12,263 (9.46\textperthousand) & 170.59 \\
            \scoopTask  & 55,784,462 & 11,363 (0.20\textperthousand) & 1,129.84 \\
            \hammerTask & 30,389,168 & 10,320 (0.34\textperthousand) & 1,248.33 \\
            \bottomrule
        \end{tabular}
    }
\end{table}

\begin{table}[!t]
\centering
\footnotesize
\setlength{\tabcolsep}{3pt}
\caption{\textbf{\exploredMetricName results of method variations on \torqueTask.}}
\label{tab:search_ablation}

\subfloat[Ablations on calibration and lookahead in \bassName{}. Search limit is set to 40,000 structures.\label{tab:search_ablation_calib}]{
\begin{tabular}{cccc}
\toprule
\multirow{2}{*}{\textbf{Lookahead step(s)}} &
\multicolumn{3}{c}{\textbf{\exploredMetricName ($\downarrow$) with calibration budget}} \\
\cmidrule(lr){2-4}
& \textbf{9,800} & \textbf{980} & \textbf{0} \\
\midrule
$N=2$ (lookahead) & 12,263 & 11,103 & $>$40,000 \\
$N=1$ (greedy) & 12,320 & $>$40,000 & $>$40,000 \\
\bottomrule
\end{tabular}
}
\\
\subfloat[\bassName{} \vs random sampling.\label{tab:search_ablation_seed}]{
\begin{tabular}{ccc}
\toprule
\textbf{Method} &
\textbf{\exploredMetricName{} ($\downarrow$) Median} &
\textbf{Range (min--max)}\\
\midrule
\bassName{} &
12,263 & 10,658--19,951 \\
Random &
766,123 & 426,699--1,202,597   \\
\bottomrule
\end{tabular}
}

\end{table}

\subsection{Task and Experiment Settings}\label{sec:tasks}

We design four tool-use manipulation tasks with distinct physical effects and structural requirements. For each task, we design $\mathcal{L}_\tau$ from the desired interaction outcomes following~\cref{sec:lower}.
\textbf{(a) \sweepTask:} The tool must approach two balls and move them into a dustpan; success requires both balls lying in the dustpan (\cref{fig:sim}(a)).
\textbf{(b) \torqueTask:} The tool must engage a bolt and rotate it by a target angle; success requires sufficient bolt rotation and a final tool pose within tolerance (\cref{fig:sim}(b)).
\textbf{(c) \scoopTask:} The tool must approach two floating balls and lift them; success requires both balls to be raised above a given height (\cref{fig:sim}(c)).
\textbf{(d) \hammerTask:} A compositional tool must first hammer a nail then pry out a second nail; success requires both nails reaching their desired depth or height (\cref{fig:sim}(d)). Due to the challenge of designing a dual-task tool, we reduce its asset library (panel excluded) for practicality.

\begin{figure}[!htb]
    \centering
    \includegraphics[width=\linewidth]{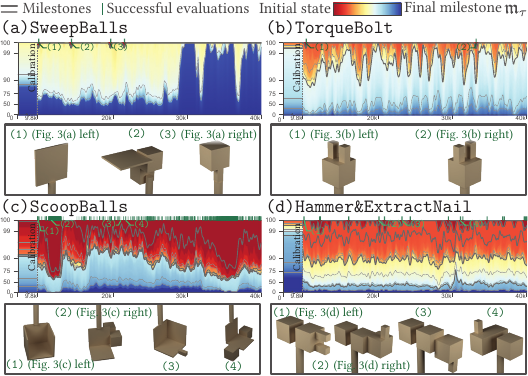}
    \caption{\textbf{Milestone achievement during \bassName{} structural search across the four tasks.} Each curve shows the rolling fraction of evaluated structures that reach a given milestone level, computed over a window of 100 consecutive evaluations. The first 9,800 evaluations constitute the random calibration phase, after which \bassName{} uses milestone feedback to guide structural search. Green ticks mark evaluations that achieve all milestones.}
    \label{fig:milestones}
\end{figure}

We run \bassName{} on each task using 140 CPU cores (Intel Xeon Gold 6248 @ 2.50\,GHz). We first perform 9,800 random calibration evaluations to estimate the prior (\cref{sec:feedback}), then use the calibrated behavioral model with prior strength $\kappa_\ell=2$ to guide structural search. Stage~1 discovers diverse successful structures on each task, including \emph{conventional} tools such as the spoon-like design for \textbf{\texttt{ScoopBalls}} and the claw-hammer-like design for \textbf{\texttt{Hammer\&ExtractNail}}, as well as unconventional \emph{creative} structures with distinct geometries and tool-use strategies; see representatives in~\cref{fig:sim}. We analyze the search behavior of \bassName{} and ablate the design choices of calibration and lookahead in~\cref{sec:exp:bass}.

Stage~1 yields a large pool of successful candidate structures, from which we selectively refine the more robust designs in Stage~2. To identify such candidates, we reconstruct each successful Stage~1 design and task scene in MuJoCo, replay its optimized SE(3) trajectory, and retain the designs with the highest cross-simulator success rates for subsequent refinement. Representative results are analyzed in~\cref{sec:exp:refine}, then deployed on a physical robot for performing tool-use tasks in the real world, as reported in~\cref{sec:real}.

\subsection{Tool Structural Search with \bassName{}}
\label{sec:exp:bass}

\bassName{} concentrates evaluations on promising structures through behavioral feedback, enabling successful tool discovery with less exploration, visualized in~\cref{fig:milestones}. For each milestone level $\ell$, we compute the rolling fraction of structures satisfying $C\geq\ell$ over a 100-evaluation window. Across tasks, post-calibration search allocates substantially more evaluations to later milestones, with high-progress regions repeatedly emerging. Green ticks denote the discovery of tool structures that successfully achieve the task, with representatives visualized in boxes below.

\cref{tab:stage1} further quantifies this efficiency. We report the \exploredMetricFullName (\exploredMetricName) that counts terminal structures evaluated from calibration through the first success, together with their fraction among all \ac{dag} terminal states. Across tasks, the first success requires exploring only $0.20$--$9.46$\textperthousand\ of the \ac{dag} terminal space: $<1\%$ in every case and $<0.4$\textperthousand\ for the two larger spaces. This shows that behavioral feedback concentrates evaluations on promising structural regions while exploring only a small fraction of the search space.

\paragraph*{Ablations}
We ablate the calibration and lookahead designs in \bassName{} using \exploredMetricName. Since lower-level optimization is computationally expensive, we conduct ablations on \torqueTask, which has the lowest evaluation cost among our tasks. \cref{tab:search_ablation_calib} compares calibration budgets and lookahead steps $N$. With full calibration, greedy search ($N=1$) performs similarly to two-step lookahead ($N=2$); as calibration is reduced, however, lookahead becomes increasingly advantageous. This suggests complementary roles for calibration and lookahead: calibration provides informative behavioral priors, while lookahead remains effective even when calibration information is limited, enabling efficient structural search from sparse evidence. We thus use full calibration in further experiments to reduce sensitivity to sparse calibration. Across multiple seeds, \cref{tab:search_ablation_seed} shows that \bassName{} reaches the first success substantially earlier and within a much narrower range than uniform random terminal sampling, demonstrating both efficient and consistent structural search.

\subsection{Tool Shape-action Refinement}
\label{sec:exp:refine}

Among the retained structures from Stage~1, Stage~2 produces functionally interpretable, task-relevant refinements when necessary, exploiting the remaining optimization headroom in shape and action.
We report the deformation ratio (DR) as the RMS of non-anchor cage vertex displacements from Stage~1, each normalized by its primitive's initial longest bounding-box side in local coordinates, aggregated over the tool or an individual primitive.

On \sweepTask, the first example (\cref{fig:sim}(a), left) is a \emph{conventional} sweeping panel that expands by 13.2\% in functional volume, reducing the task loss by 13.4\% while both balls remain successfully delivered to the target region. The second example (\cref{fig:sim}(a), right) shows a \emph{creative} cuboid--cube structure similarly concentrating deformation on the task-relevant working cube (6.94\% DR \vs 1.39\% for the connector), increasing the terminal safe margin from 4.12~mm to 9.20~mm while preserving success. On \torqueTask, both examples show \emph{conventional} wrench-like structures. The first example (\cref{fig:sim}(b), left) has a pair of jaws whose inner opening tightens and develops a taper along depth. For the 6-mm-wide bolt head, the static clearance decreases from 1.07~mm to 0.49~mm at the jaw root ($-54.2\%$), but only to 0.78~mm near the entrance ($-27.1\%$), forming a lead-in that remains loose for alignment and tightens deeper inside to reduce wobble. Correspondingly, the handle-bolt rotation mismatch decreases from $11.97^\circ$ to $6.95^\circ$ ($-41.9\%$), while the two working jaws deform 2.4--2.7$\times$ more than the central connector, despite no part-level supervision. The second example (\cref{fig:sim}(b), right) shows the complementary regime: with only 0.23\% DR, Stage~2 reduces the loss by 9.1\% and the rotation mismatch from $12.13^\circ$ to $6.98^\circ$, showing that substantial improvement can also arise from Stage~2 action refinement. \scoopTask provides a clear contrast between the two structures. The \emph{conventional} spoon-like example (\cref{fig:sim}(c), left) changes only slightly (0.75\% DR and 0.2\% loss reduction), indicating that its Stage~1 geometry already suits the task and is therefore largely preserved. The other (\cref{fig:sim}(c), right) admits substantial beneficial refinement of its asymmetric structure: its bottom panel becomes 87.8\% thinner, while the tool-head mass decreases by 33.7\%; together with action re-optimization, this yields a 46.2\% loss reduction while preserving successful ball retention through the lift. On \hammerTask, both examples develop task-specific geometries at both functional ends. In the \emph{conventional} example, the two claw leaves become 13.4--17.7\% thinner along depth, with larger thinning near the root than at the tip, while the hammer block expands by 4.7\% and 4.4\% along two bounding directions. These lead to a 1.7\% increase in nail-down depth during hammering, illustrating that Stage~2 selectively reshapes different functional regions according to their roles.

These results show that Stage~2 does not impose a fixed mode of geometric refinement. Depending on the Stage~1 structure, it can substantially reshape task-relevant geometry, make localized adjustments, primarily refine the action with minimal deformation, or preserve an already effective shape. This provides a physically interpretable continuous refinement of the structures discovered in Stage~1.

\subsection{Real-World Tool Use}\label{sec:real}

\begin{figure}[t!]
    \centering
    \includegraphics[width=\linewidth]{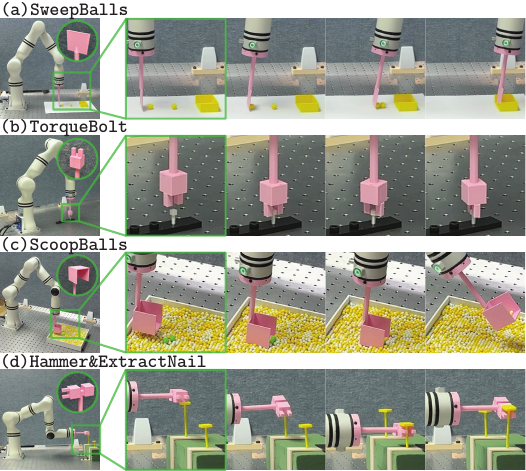}
    \caption{\textbf{Real-world tool use.} The tools are 3D printed, mounted on a RealMan RM75-B arm, which executes the rolled-out trajectories.}
    \label{fig:real}
\end{figure}

We 3D print the final optimized geometries of the representative tools, mount them on a 7-DOF RealMan RM75-B arm, and replay the actions recorded from successful simulation rollouts. As \cref{fig:real} shows, the tools complete all four tasks in the real world. Under identical initial conditions, the refined tools behave more favorably than their Stage~1 counterparts: in \torqueTask, the tighter jaw opening fits the bolt head with less wobble for more stable engagement, and in \sweepTask, the enlarged panel covers a wider region per stroke, controlling the ball trajectories more tightly and reducing escapes. Overall, the real-world experiments confirm the physical usability of the design pipeline and the sim-to-real reproduction of the optimized tool-use behaviors.

\section{Conclusions}

We presented \frameworkName{}, a task-driven framework that designs robot tools from scratch by determining structure, shape, and action jointly from the desired physical outcome. Tool design is cast as a bilevel optimization and solved by a two-stage approximation, with \bassName{} guiding the structural search using the milestone progress of evaluated tools as behavioral evidence. On four tasks, \frameworkName{} discovers functional structures without prescribed tool templates and refines their shapes and actions to lower the task loss while preserving success, and the designed tools accomplish the tasks on a real robot.

Several directions remain. First, GPU-parallel differentiable simulators would permit larger search budgets and faster optimization. Second, injecting randomization for task settings may lead to tools with more robust designs. Third, a richer asset library would extend the framework to more complex tools and thus more complex functionalities.

\section*{Acknowledgement}

This work is supported in part by the Brain Science and Brain-like Intelligence Technology---National Science and Technology Major Project (2025ZD0219400), the Beijing Natural Science Foundation (QY26049),  the National Natural Science Foundation of China (62376009), the Beijing Nova program, the NVIDIA Academic Grant Program using Spark and Thor, the State Key Lab of General AI at Peking University, the PKU-BingJi Joint Laboratory for Artificial Intelligence, the Wuhan Major Scientific and Technological Special Program (2025060902020304), the Hubei Embodied Intelligence Foundation Model Research and Development Program, and the National Comprehensive Experimental Base for Governance of Intelligent Society, Wuhan East Lake High-Tech Development Zone. We thank Ms. Hailu Yang (PKU), Ms. Qing Gao (PKU), and Lulin Huang (PKU) for their assistance.

\bibliographystyle{IEEEtran}
\bibliography{reference_header_shorter,reference}

\end{document}